# Generating Weather Forecast Texts with Case based Reasoning

Ibrahim Adeyanju
IDEAS Research Institute
The Robert Gordon University
Aberdeen, Scotland, UK

## ABSTRACT

Several techniques have been used to generate weather forecast texts. In this paper, case based reasoning (CBR) is proposed for weather forecast text generation because similar weather conditions occur over time and should have similar forecast texts. CBR-METEO, a system for generating weather forecast texts was developed using a generic framework (jCOLIBRI) which provides modules for the standard components of the CBR architecture. The advantage in a CBR approach is that systems can be built in minimal time with far less human effort after initial consultation with experts. The approach depends heavily on the goodness of the retrieval and revision components of the CBR process. We evaluated CBR-METEO with NIST, an automated metric which has been shown to correlate well with human judgements for this domain. The system shows comparable performance with other NLG systems that perform the same task.

## 1. INTRODUCTION

In applied Natural Language Generation (NLG), the domain of weather forecasting is very popular and has been used to test the effectiveness of several text generation techniques [1,2,3,4,5,6,7]. Many techniques have been proposed and applied to automated generation of weather forecast texts. Such techniques include knowledge intensive approaches in which explicit rules are elicited from domain experts and corpus analysis [6,8] at different stages of the text generation process. Machine learning models, especially statistical methods, have also been used to design systems that learn generation models introspectively from the corpus [7,9]. The use of machine learning to build text generation models is knowledge-light.

However, weather forecasting and generation of weather forecast texts are natural Case Based Reasoning (CBR) problems. This is because the basic assumptions in CBR are that *similar problems occur again* and *similar problems have similar solutions*. When we look at the weather conditions for a particular day, we are immediately reminded of similar weather conditions in the past. Therefore, it is expected that similar weather conditions should have similar forecast texts and it is easier to reuse previous similar forecast texts to generate new forecast texts. In this paper, we show how CBR as a knowledge-light approach can be used to generate weather forecast texts from forecast data and discuss its merits and demerits.

Background and related works appear in Section 2 followed by a description of the CBR architecture and its application to text generation in Section 3. Section 4 discusses our experimental setup and evaluation results while our conclusion appears in Section 5.

## 2. BACKGROUND

Automated generation of weather forecast texts has been achieved using several techniques. These techniques can be divided into two broad categories: knowledge-intensive (KI) and knowledge-light (KL) approaches. KI approaches require extensive consultation with domain experts during corpus analysis and throughout the text generation process. On the other hand, KL approaches rely more on the use of automated methods which are mainly statistical.

One of the earliest KI systems generated forecast texts by inserting numeric values in standard manually-created templates [10]. Multiple templates are created for each possible scenario and one of them is randomly selected during text generation to provide variety. Other KI systems such as ICWF [11], FoG [12] and SumTime [6] developed linguistic models using manually-authored rules obtained from domain experts and corpus analysis. Some of these systems, e.g. FoG and SumTime, used NLG architecture [13] where the generation process is separated into different modules. The modules in architecture include document planning, micro planning and realization.

The KL approach to generate forecast texts typically employs machine learning techniques. Trainable systems are built using models based on statistical methods such as probabilistic context-free grammars and phrase based machine translation [14]. The advantage is that systems are built in less time and with less human effort as compared to the KI approach. Forecast texts generated by KL systems were reported to have comparable quality to KI systems when evaluated with automated metrics [15]. However, KI systems were better when evaluated by humans.

Synergy between CBR and NLG has previously been exploited for automatic story plot generation [16,17]. Here, a plot structure is obtained by reusing stories from a case base of tales and ontology of explicitly declared relevant knowledge. NLG techniques are then used to describe the story plot in natural language. Although the story generated is a complete sketch of the plot, it assists screen writers in fast prototyping of story plots which can easily be developed into a story.

## 3. CBR APPROACH TO GENERATING WEATHER FORECAST TEXTS

This section gives an insight into the case based reasoning (CBR) paradigm and how it is used for generating weather forecast texts. The basic concepts and terminology in CBR are discussed using examples from the weather forecast domain.

### 3.1 Case Based Reasoning

The basic principles in CBR are that similar problems occur again and similar problems have similar solutions. It is therefore easier to modify a previous solution to a similar problem than solving a new problem from scratch. The technique therefore requires knowledge in the form of problem-solving episodes where each episode is called a *case*. Each case consists of a problem and its solution and a set of cases form the casebase. The CBR problem-solving architecture as shown in Figure 1 typically consists of four components: Retrieve, Reuse, Revise and Retain commonly referred to as 4Rs [18].

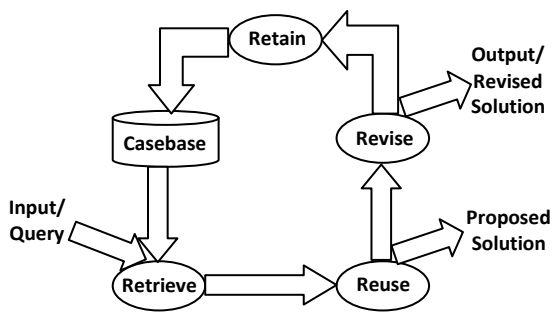

**Figure 1: A typical CBR architecture**

The most similar case/cases is/are *retrieved* from the casebase when a new problem (denoted as input or query) is encountered. The information and knowledge in the retrieved similar case/cases is/are then *reused* to solve the current problem. Typically, formulating the proposed solution might require modification to the retrieved solution to compensate for problem mismatches if the retrieved similar case is not an exact match. Adaptation, which is a sub-component of case reuse, takes into account the differences between the problems (input and retrieved) to guide any adjustment required in the proposed solution. The revision component ensures that a proposed solution is evaluated for its accuracy by a human or generic domain models/rules. The proposed solution is *revised* by such human expert if it does not accurately solve the problem at hand. Finally, a new case consisting of the input (problem query) and output (revised solution) is reviewed and *retained* while maintaining an efficient case-base by excluding redundant or noisy cases.

### 3.2 CBR-METEO: A weather forecast text generation system

Our system, which we call CBR-METEO, generates weather forecast texts using a repository of previous forecast texts available in a casebase. In this domain, a case consists of a pair of weather attribute values (for parameters like humidity, out-look, wind speeds, wind directions and forecast times) and equivalent forecast texts generated by human experts. We restrict our weather data to those related to wind forecasts (i.e. wind speeds, directions, gusts and time) and associated texts for simplicity. CBR-METEO therefore generates a wind forecast text for a new wind input data but can be easily extended for other weather parameters. The system is built using jCOLIBRI [19], an existing CBR framework. jCOLIBRI provides generic modules for each component of the typical CBR architecture (see Figure 1) thereby making development of systems easier and faster.

The input or query to CBR-METEO consists of a number of weather forecast states during the period (usually a day) in which a forecast text is required. Table 1 shows an example of the input to our system for wind weather. An input typically has two, three or four states and each wind state has values for the following attributes: wind direction (Wind Dir), minimum wind speed (Low Spd) and maximum wind speed (High Spd) and a time stamp (Time) indicating for what time of the day the data is valid. Also, minimum gust speed (Lw Gst) and maximum gust speed (Hg Gst) sometimes appear in input wind states. The speed values for wind and gust are measured in nautical miles (Knots). A dash ('-') is used to show that the value of a particular weather attribute is absent.

**Table 1: Sample input to CBR-METEO**

| State | Time | Wind Dir | Low Spd | High Spd | Lw Gst | Hg Gst |
|---|---|---|---|---|---|---|
| 1 | 0600 | n | 6 | 10 | - | - |
| 2 | 2400 | ne | 15 | 20 | - | - |

n 10 or less gradually veering ne 15-20

**Figure 2: Sample output from CBR-METEO**

CBR-METEO's output is the forecast text generated from the input data. Figure 2 shows an example of a wind forecast text generated by the system for the input data in Table 1. Other components of the system are discussed below.

### 3.3 Retrieval

The text generation process in CBR-METEO begins with the retrieval of a case from the casebase whose weather data is most similar to the input (query). Defining how the similarity metric is therefore very important for the retrieval component. The best form of similarity minimizes the work done by the succeeding components of reuse and revise.

Our similarity computation ensures that a retrieved similar weather data must have the same number of states as the input. This is because the number of states usually determines the number of phrases in the forecast text. Time attributes are compared using the differences between time stamps in aligned wind states. We then define similarity between weather data (i.e. input and each case in the casebase) mainly in terms of patterns across wind states. The patterns for a scalar attribute (e.g. wind speed) are *increasing*, *decreasing* or *constant* as we move from one state to another while *veering* (clockwise), *backing* (anti-clockwise) or *stable* patterns are applicable to vectorial attributes (e.g. wind direction). The input and weather data in each case in the casebase are transformed into a representation showing the pattern transition across wind states for each scalar and vectorial attribute.

Retrieval is done in a step-wise or hierarchical manner in which all previous cases having the same number of wind states as the input data are first retrieved. Cases within this retrieval set whose average time stamp differences from the input are within a specified threshold are then selected. Similarly, cases with the same weather patterns (for wind speeds, directions and gusts) as the input across states are then selected iteratively from preceding subset of cases. If more than one case is retrieved at the end of this iterative process, the most similar case is chosen as the one with the shortest average distance between its wind states and the input's. The distance between wind states is computed by first converting wind directions into their numeric angular values (in degrees). Each pair of wind speed and angular direction is taken as a vector quantity which represents a wind state. Cosine rule is then used to compute the distance between vectors and an average taken across the number of wind states in the input. However, if no case is retrieved as the end of the iterative process, the system gives no forecast text. The stepwise retrieval ensures that retrieved cases are semantically similar to the input data and therefore minimal modifications are carried out by the reuse and revision components.

**Table 2: Retrieved case similar to input data**

| State | Time | Wind Dir | Low Spd | High Spd | Lw Gst | Hg Gst |
|---|---|---|---|---|---|---|
| 1 | 0600 | nnw | 8 | 10 | - | - |
| 2 | 2400 | nne | 13 | 15 | - | - |
| **Forecast Text** | | | | | | |
| nw-nnw 8-13 gradually veering nne 10-15 | | | | | | |

Table 2 shows the best case retrieved for the input data in Table 1. The retrieved case is most similar to the query not only because they have the same number of states and time stamps but the wind speed and direction patterns are also similar. The wind direction as we move from state 1 to 2 in both the input data ($n \rightarrow ne$) and retrieved case ($nnw \rightarrow nne$) are veering (i.e. clockwise). The increasing wind speed pattern is also common to both; an average speed of 8/9 knots in the early morning (6a.m.) to 17/14 knots by midnight in the input/retrieved case respectively. An example of how to compute the distance between two wind states using cosine rule as the last step in retrieval process is shown in Figure 3. The values shown in the example are those from the first wind states for the input and the retrieved similar wind data. The wind speed shown for each wind state is an average of the minimum and maximum wind speeds.

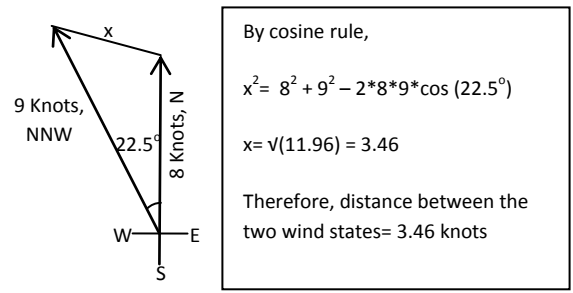

**Figure 3: Computing distance between two wind states**

## 3.4 Reuse
The reuse component of CBR-METEO puts the forecast text associated with the retrieved similar wind data in the context of the input. To do this, the forecast text is parsed to identify attribute values from the retrieved wind data that are present in the text. These attribute values are then substituted with their equivalent from the input. No action is carried out by the reuse component if the retrieved wind data is identical to the input. In other words, the forecast text associated with retrieved wind data can be returned directly for output if the similarity between the input and retrieved data equals 1 at every step during the hierarchical computation of similarity.

n 6-10 gradually veering ne 15-20

**Figure 4: Sample forecast text by CBR-METEO**

An example of the forecast text from the reuse phase is shown in Figure 4 with the input and retrieved forecast text in Tables 1 and 2 respectively. Here, the wind speeds (8-13 & 10-15) and directions (*nw-nnw* & *nne*) in the retrieved text are replaced with those from the input.

## 3.5 Revision
The revision component uses expert rules to ensure that specific phrases conform to writing conventions in the domain. Such rules are learnt during post-edit tasks where experts are given input data and forecast texts proposed by the reuse component. Figure 5 shows a revised form of the forecast text in Figure 4 where one of expert rules is applied to revise "*6-10*" in the reuse forecast text into "*10 or less*".

n 10 or less gradually veering ne 15-20

**Figure 5: An example of a revised forecast text**

## 3.6 Retain
Retention can be carried out in CBR-METEO where new cases consisting of the input and output (generated forecast texts) are be added to the casebase after further review by experts. Inputs whose forecast texts CBR-METEO was unable to generate can also be added after generation by the experts or using other techniques. The system thereby evolves over a period of time and is able to generate accurate forecast texts for most inputs (if not all) when this component is functional.

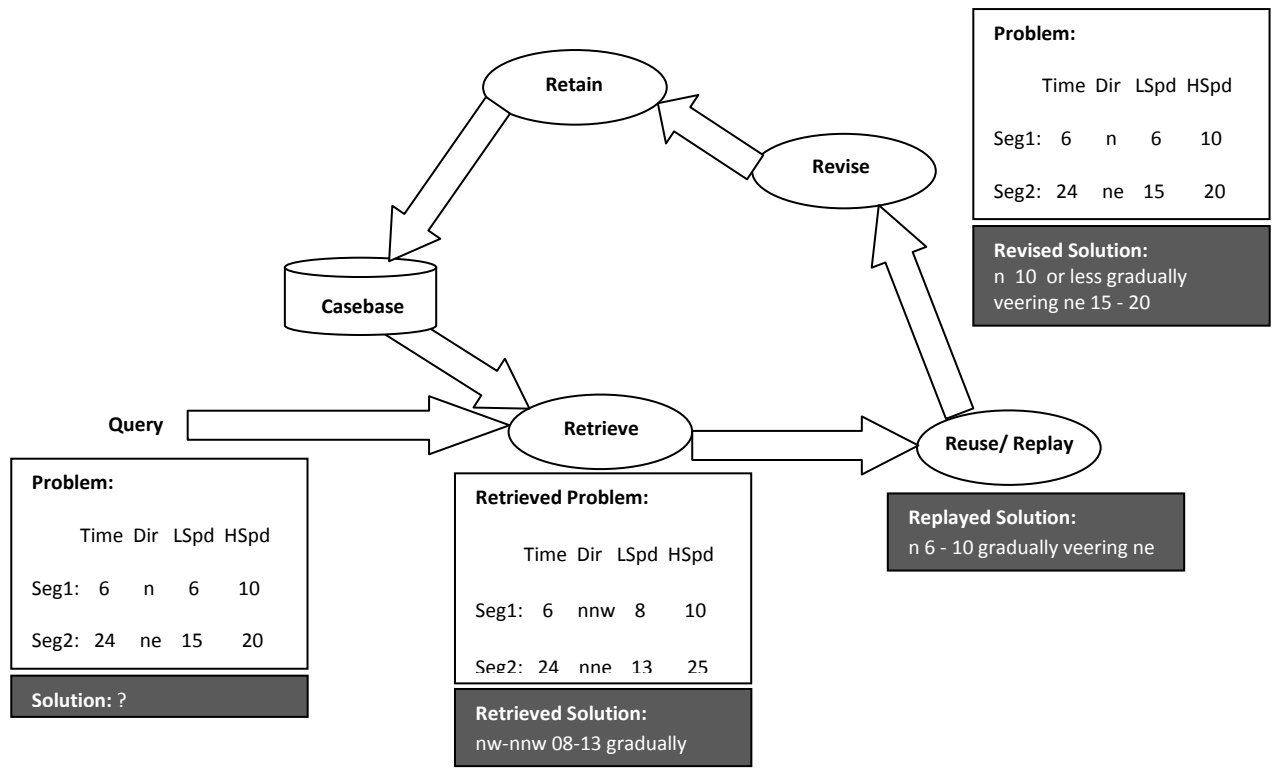

**Figure 6: Generating wind forecast texts with CBR-METEO**

## 3.7 CBR-METEO: A guided illustration

A summary of the text generation process is illustrated in Figure 6. The figure shows the transition between the different components (retrieve, reuse, revise and retain) with the same examples used in Sections 3.2 to 3.6. Here, the wind data is the problem and the forecast text is solution as viewed in the CBR context. The wind attributes shown are wind direction (Dir), minimum wind speed (LSpd) and maximum wind speed (HSpd). The attributes associated with gusts are not shown because they are absent in the input used for illustration.

## 4. EXPERIMENTAL SETUP

We employed a five hold-out experimental design identical to [14] in our experiments. We evaluate the forecast text generated by CBR-METEO with NIST [20] which correlates best with expert judgements in the domain of weather forecast text generation [15] as compare to other automated evaluation methods such as BLEU [21] and ROUGE [22]. We then compare the results with ten existing NLG systems; Sum-Time hybrid [15] and nine trainable systems [7].

### 4.1 Dataset

Our experiments were evaluated using the wind weather corpus described in [23]. The corpus consists of wind forecast data and texts divided into five folds where each fold is further sub-divided into training and test sets. The training has 2104 wind forecasts while there are 221 forecasts in the test set across all five fold with duplicates. The test sets were used as our queries while each corresponding training set is the casebase in our experiments. The wind data was parsed from the human-authored forecast text where each phrase produces a vector of 7-tuples *(i, d, smin, smax, gmin, gmax, t)* where *i* is the tuples ID, *d* is the wind direction, *smin* and *smax* are the minimum and maximum wind speeds, *gmin* and *gmax* are the minimum and maximum gust speeds, and *t* is a time stamp (indicating for what time of the day the data is valid). If one or more parts of the 7-tuple is not realised in a given forecast, a '−1' value is shown for a timestamp and a '-' value for the speed, gust or direction attribute.

The forecast texts consist of natural language forecasts from human forecasters, Sum-Time hybrid system [15] and nine trainable systems [7]. The nine trainable systems include 5 probabilistic context free grammar (PCFG) systems, 2 probabilistic synchronous context free grammar (PSCFG) systems and 2 phrase-based statistical machine translation (PBSMT) systems. The major difference in the trainable systems is their mode of generation; PCFG system has five modes: greedy (PCFG-greedy), roulette (PCFG-roule), viterbi (PCFG-viterbi), n-gram (PCFG-2gram) and random (PCFG-rand). Likewise, the PSCFG system modes are semantic (PSCFG-sem) and unstructured (PSCFG-unstr) while PBSMT system has unstructured (PBSMT-unstr) and structured (PBSMT-struc) modes.

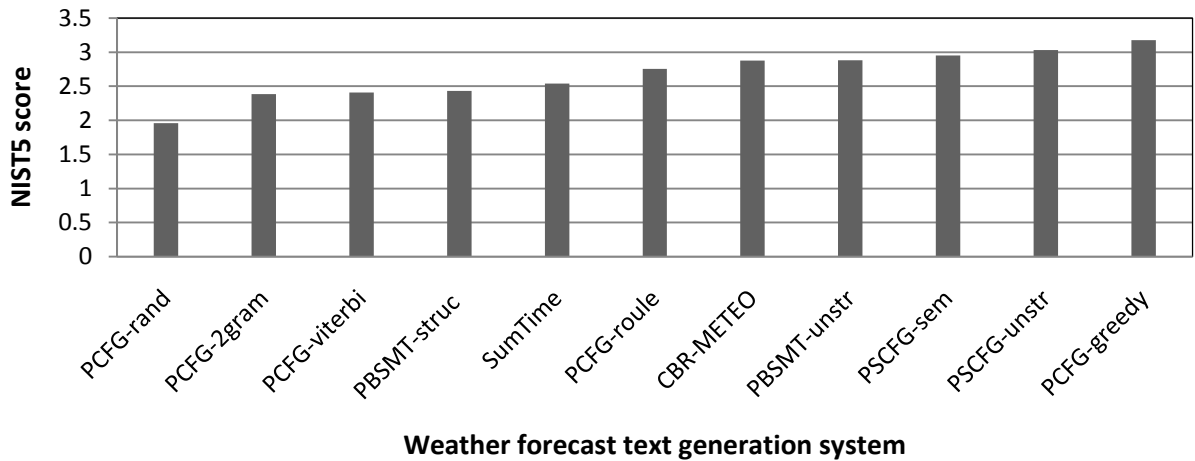

Figure 7: Graph of average NIST5 scores for CBR-METEO and 10 other NLG systems

## 4.2 Evaluation results

Figure 7 shows a graph of average NIST5 scores from experiments on the wind forecast corpus with the CBR and NLG systems. The effectiveness of the CBR system is comparable to Sum-Time as well as the trainable NLG systems as shown on the graph. Although, four trainable systems (PBSMT-unstructured, PSCFG-semantic, PSCFG-unstructured and PCFG-greedy) outperform our CBR system according to the results, its performance is very similar to the SumTime hybrid system which was ranked as best by human evaluators [14]. This indicates that the performance might also be ranked high if it were evaluated by humans. The minimal hand-coded rules used for revision of the replayed forecast text in CBR-METEO were similar to SumTime's as they are obtained from domain experts and this might account for the comparable performance.

The results shown in Figure 7 are averages across the number of forecast texts generated by CBR-METEO or available from the corpus for the other ten NLG systems. CBR-METEO is only able to generate forecast texts for 137 out of 221 inputs (queries) in our test sets. This is because it generates no forecast text when the casebase does not contain a case with the same wind patterns (for speed and direction) as the input. The system was designed this way since matching similar wind patterns (rather than exact patterns) requires the application of more complex revision rules after the reuse phase. However, the revision component of CBR-METEO uses only minimal rules to correct the forecast text from the reuse phase. Such minimal rules enable simple revisions such as changing the speed in the forecast text from a number range into a phrase when it is less than 10. There is a trade-off between the similarity definition that determines the number of inputs for which CBR-METEO can generate text and the complexity of revision rules. While retrieval of similar wind patterns allows the system to generate text for more queries, complex expert rules such as the ones used in SumTime will be required at the revision stage rather than minimal revision rules.

We made a number of observations during our experiments which gave us an insight to why the NIST evaluation rank the performance of CBR- METEO lower than some of the trainable systems though human evaluators might think otherwise. First is that the NIST scoring and trainable systems are inherently based on statistical methods; therefore the trainable systems are more likely to be ranked high. Another issue is the stylistic variation in forecaster's text for time phrases and change verbs. For example while some forecasters use 'increasing', others use 'rising' in their forecast text. The same reason applies for the use of 'decreasing' versus 'falling'. Such stylistic variations will be better captured by NIST if there are multiple human references for each test input as opposed to just one human reference used in our experiments. Another issue with the human reference that might affect the evaluation results is the use of wind sub-directions such as 'nw-nnw' or 'ne-e' by forecasters though the forecast data contains 'nnw' or 'nne'. Such change to the original forecast data during writing of the text is not uncommon as it makes the forecast text more reliable but it is not used consistently among forecasters. The ellipsis of the time phrase where forecast time phrases are implied in the forecast text can also affect the automated evaluation measure. For example, some forecasters add 'by late evening' to the forecast phrase for time stamp '2100' while others don't especially if it is the last wind state to be written in the forecast text. The observations (i.e. stylistic variation, inconsistent usage of wind sub-directions and time phrase ellipsis) adversely affect the automated evaluation of CBR-METEO since CBR-METEO does not take into account the profile of different human forecasters during retrieval. The errors associated with these observations might be minimized by incorporating knowledge about authors' of human forecast texts; however, this was not available in the dataset.

In order to improve the performance of our system further, we need to allow more ambitious revisions than the current system. This means our retrieval component needs to be less strict while matching cases. For example, a query with two time stamps 0600 and 1800 can be matched to a retrieved case with 0600 and 2100 but the revision stage will have to change phrase "late evening" to "evening" to account for the fuzzy match. The fuzzy similarity is also applicable when matching wind speeds and directions. However, the fuzzier the matching at the retrieval stage, the more revision rules that will be required to get the final solution text. This is obvious since the revision component is dependent on retrieval and reuse. When fuzzy matching is allowed, the case similarity value needs to be propagated to the revision component and used to determine the amount of modification required to obtain an accurate output. We are currently working on improving our revision component. One specific direction we are working on is to use another casebase usually called adaptation casebase [24] to store cases of text revisions made by domain experts. The adaptation casebase will be similar to the post-edit corpus currently available in the domain [25].

## 5. CONCLUSION

In this paper, we presented an approach to generating weather forecast texts using CBR technology. This involves the retrieval of previous similar weather data in response to an input data whose text is required. The forecast text associated with the retrieved similar weather data is then reused in the context of the new data followed by minimal revision. The approach is knowledge-light therefore ensuring that systems can be built in little time and with less human effort. CBR also allows for a gradual evolution of the system since new forecast texts can retained for future use. Our system, CBR- METEO, was evaluated against other NLG systems performing the same task and showed comparable results.

The limitation of our current system is that it cannot generate forecast texts for all queries if a previous similar weather data is not found in the casebase. We intend to improve on this by relaxing our similarity constraints to retrieve less similar cases. However, this will require the use of more complex rules during revision. We also intend to carry out a qualitative evaluation for our system and apply the CBR approach to other NLG tasks. Our long term goal is to study synergies between NLG and CBR techniques and apply them to develop better and more effective AI systems.

## 6. ACKNOWLEDGMENTS

The author will like to thank Nirmalie Wiratunga of The Robert Gordon University, UK and Somayajulu Sripada of University of Aberdeen, UK for their useful feedback.

## 7. REFERENCES

[1] Kittredge, R., Polguére, A. and Goldberg, E. 1986. Synthesizing weather reports from formatted data. In *Proceedings of the 11th. International Conference on Computational Linguistics*, 563–565.

[2] Bourbeau, L., Carcagno, D., Goldberg, E., Kittredge, R., and Polgure, A. 1990. Bilingual generation of weather forecasts in an operations environment. In *Proceedings of COLING'90*, 318–320.

[3] Sigurd, B., Willners, C., Eeg-Olofsson, M., and Johansson, C. 1992. Deep comprehension, generation and translation of weather forecasts (weathra). In *COLING-92*, 749–755.

[4] Coch, J. 1998. Interactive generation and knowledge administration in multimeteo. In *Proceedings of the 9th International Workshop on NLG*, 300–303.

[5] Rubinoff, R. 2000. Integrating text planning and linguistic choice without abandoning modularity: the IGEN generator. *Computational Linguistics*, 26(2):107–138.

[6] Sripada, S., Reiter, E. and Davy, I. 2003. SumTime-Mousam: Configurable marine weather forecast generator. *Expert Update*, 6(3):4–1.

[7] Belz, A. 2007. Automatic generation of weather forecast texts using comprehensive probabilistic generation-space models. *Natural Language Engineering*, 14:431–455.

[8] Reiter, E., Sripada, S. and Robertson, R. 2003. Acquiring correct knowledge for natural language generation. *Journal of Artificial Intelligence Research*, 18:491–516.

[9] Dimitromanolaki, A. and Androutsopoulos, I. 2003. Learning to order facts for discourse planning in natural language generation. In Proc. of *EACL Workshop on NLG*.

[10] Glahn, H. 1970. Computer-produced worded forecasts. *American Meteorological Society Bulletin*, 51(12):1126–1131.

[11] Ruth, D.P. and Peroutka, M.R. 1993. The interactive computer worded forecast. In *9th International Conference on Interactive Information and Processing Systems for Meteorology, Oceanography and Hydrology*, 321–326. American Meteorological Society.

[12] Goldberg, E., Driedger, N. and Kittredge, R. 1994. Using natural-language processing to produce weather reports. *IEEE Expert*, 9:45–53.

[13] Reiter, E. and Robert Dale. 1995. Building applied natural language generation systems. *Natural Language Engineering*, 1:1–32.

[14] Belz, A. and Kow, E. 2009. System building cost vs. output quality in data-to-text generation. In *Proceedings of 12th European Workshop on NLG*.

[15] Belz, A. and Reiter, E. 2006. Comparing automatic and human evaluation of NLG systems. In *Proceedings of EACL'06*, 313–320.

[16] Gervás, P. 2001. Automatic Generation of Poetry using a CBR Approach. In *Proceedings of the Conference of the Spanish Association for Artificial Intelligence (CAEPIA)*.

[17] Gervás, P., Díaz-Agudo, B., Peinado, F. and Hervás, R. 2004. Story plot generation based on CBR. In *12th Conference on Applications and Innovations in Intelligent Systems*.

[18] Aamodt, A. and Plaza, E. 1994. Case-based reasoning: Foundational issues, methodological variations and system approaches. *Artificial Intelligence Communications (AICom)*, 7(1):39–59.

[19] Díaz-Agudo, B., Gonzalez-Calero, P.A., Recio-García, J.A. and Sanchez, A. 2007. Building CBR systems with jCOLIBRI. *Special Issue on Experimental Software and Toolkits of the Journal Science of Computer Programming*, 69(1-3):68–75.

[20] Doddington, G. 2002. Automatic evaluation of machine translation quality using n-gram co-occurrence statistics. In *Proceedings of ARPA Work- shop on Human Language Technology*.

[21] Papineni, K., Roukos, S., Ward, T. and Zhu, W-J. 2002. BLEU: A method for automatic evaluation of machine translation. In *Proceedings of the 40th Annual Meeting of the Association for Computational Linguistics*, 311–318.

[22] Lin, C-Y. and Hovy, E. 2003. Automatic evaluation of summaries using n-gram co-occurrence statistics. In *Proceedings of the Human Technology Conference (HLT-NAACL 03)*, 71–78.

[23] Belz, A. 2009. *Prodigy-METEO: Pre-Alpha Release Notes*. University of Brighton, UK, first edition.

[24] Craw, S., Wiratunga, N. and Rowe, R.C. 2006. Learning adaptation knowledge to improve case-based reasoning. *Artificial Intelligence*, 170:1175–1192.

[25] Sripada, S., Reiter, E., Hunter, J. and Yu, J. 2002. SUMTIME-METEO: Parallel corpus of naturally occurring forecast texts and weather data. Technical report, Department of Computer Science, University of Aberdeen.